\documentclass[11pt]{article}
\usepackage[final]{acl}

\usepackage{times}
\usepackage{latexsym}
\usepackage[T1]{fontenc}
\usepackage[utf8]{inputenc}
\usepackage{microtype}
\usepackage{graphicx}
\usepackage{xcolor}
\usepackage{amsmath}
\usepackage{amssymb}
\usepackage{booktabs}
\usepackage{multirow}
\usepackage{array}
\usepackage{makecell}
\usepackage[most]{tcolorbox}
\usepackage{listings}

\definecolor{promptback}{RGB}{242,242,242}
\definecolor{promptframe}{RGB}{219,219,219}

\newtcblisting{promptbox}[2][]{enhanced,
  breakable,
  listing only,
  colback=promptback,
  colframe=promptframe,
  colbacktitle=promptframe,
  boxrule=0pt,
  arc=0pt,
  frame hidden,
  borderline west={0.8mm}{0mm}{promptframe},
  before skip=10pt,
  after skip=10pt,
  width=\columnwidth,
  left=0pt,
  right=0pt,
  top=0pt,
  bottom=0pt,
  fonttitle=\bfseries,
  coltitle=black,
  title={#2},
  listing options={
    basicstyle=\ttfamily\footnotesize,
    breaklines=true,
    breakatwhitespace=false,
    columns=fullflexible,
    keepspaces=true,
    showstringspaces=false
  },
  #1
}

\title{Enhancing Event Candidate Acquisition for Event Linking}

\author{
\textbf{Ziyang Zhang} \quad
\textbf{Yinan Liu}\thanks{Corresponding author.} \quad
\textbf{Boyi Xue} \\
\textbf{Yingxuan Huang}\quad
\textbf{Bin Wang} \quad
\textbf{Xiaochun Yang}
\\
School of Computer Science and Engineering, Northeastern University\\
Shenyang 110819, China\\
2472096@stu.neu.edu.cn, liuyinan@cse.neu.edu.cn, xueboyi@mails.neu.edu.cn,\\
huangyx6@mails.neu.edu.cn, binwang@mail.neu.edu.cn, yangxc@mail.neu.edu.cn
}

\begin{document}
\maketitle
\begin{abstract}
Event linking associates event mentions in text with entries in a knowledge base (KB), or identifies them as out-of-KB events. Although existing methods use different architectures, candidate event acquisition can still be weakened by short ambiguous mentions, noisy arguments, and evidence that is unevenly useful for retrieval. We present MACE, a Multi-Agent Candidate Event acquisition method that refines event structure before linking. MACE uses evidence-specialized LLM agents to acquire time, location, participant, and event-type evidence, exposes intermediate queries to candidate-event lookup tools, and lets a coordinator revise the evidence set before final candidate construction. Experiments on two event linking benchmarks show that adding MACE to different event linking models consistently improves accuracy. These results show that MACE improves event linking through better candidate event acquisition without modifying the event linking model.

\end{abstract}
% Main text
\section{Introduction}
Event linking is an important text disambiguation task. The objective of this task is to associate event mentions in text with corresponding knowledge base (KB) entries, or to recognize them as out-of-KB events when no appropriate entry is found \cite{yu2023event,hsu2024argument,liu2026sefel}. By leveraging the events' background knowledge from the KB, it can enhance text understanding, thereby enhancing downstream applications (e.g., question answering \cite{li2024meqa, liu2016deola, luo-etal-2025-etrqa} and fake-news mitigation \cite{wang2022veracity, liu2020named}). Meanwhile, event knowledge mentioned by texts is useful for enriching KBs \cite{liu2022personal, liu2026joint}.

Existing event linking methods can be roughly divided into three macro families. The first family follows the two-stage retrieve-and-rank framework represented by BLINK~\cite{wu2020scalable}: a bi-encoder is used for candidate event retrieval, and a cross-encoder is used for fine-grained ranking over the retrieved candidate events. EveLink~\cite{yu2023event} extends this framework with local named entities, and the argument-aware approach of Hsu et al.~\cite{hsu2024argument} further incorporates event arguments into the model input. The second family uses generation-based methods, represented by GENRE~\cite{de2020autoregressive}, which formulates event linking as autoregressive generation of the target title. The third family uses fast argument-aware representation-based methods, represented by SEFEL~\cite{liu2026sefel}, which builds argument-aware event mention representations, treats \textit{NIL} as an explicit candidate, and scores candidates in a shared embedding space without a separate reranker.

However, existing studies still leave two candidate-event acquisition issues insufficiently addressed. First, event arguments differ in how much they help disambiguate candidate events, but existing systems usually pass extracted arguments to the event linking model without explicitly selecting the most retrieval-useful ones. For example, in the sentence \textit{The 1st Infantry Division landed in Normandy on June 6, 1944}, the participant \textit{1st Infantry Division} points to a much narrower set of candidate events than the location \textit{Normandy}, which can match many Normandy-related events. Second, extracted arguments are not always verified before they are used to form candidate event queries. Unsupported or peripheral spans can therefore expand the search toward many irrelevant candidate events and weaken the final event linking decision.

To address these issues, we propose MACE, which refines event structure before event linking by extracting, verifying, and selecting typed event arguments for candidate event acquisition. MACE uses evidence-specialized LLM agents for time, location, participant, and event-type evidence, converts scattered context clues into typed arguments, filters unsupported or redundant outputs, and observes candidate-event lookup results through a bounded coordinator loop. The coordinator maintains the current evidence state and decides whether to accept, reverify, refine, drop weak evidence, or request targeted re-extraction before final candidate construction. Since MACE modifies candidate event acquisition rather than the event linking model, it can enrich the candidate event sets used by various event linking methods.

The main contributions of our work can be summarized as follows: (1) We propose MACE, which is the first plug-in for enhancing candidate event generation in event linking, built as a multi-agent pipeline that extracts, verifies, and revises event evidence before constructing retrieval inputs. (2) Experimental results show that MACE is flexible and plug-and-play across different event linking backbones, improving performance without modifying the underlying model architectures.

\section{Task Definition}

Let $\mathcal{T}=\{x_1,x_2,\ldots,x_n\}$ be the input text, let $m=\{x_i,\ldots,x_j\}\subset \mathcal{T}$ be the target event mention, and let $\mathcal{T}_m$ be its local context. Given a KB $Q$ with event entries $\mathcal{E}$, event linking predicts a label $y$ for $m$ from $\mathcal{E}\cup\{\text{NIL}\}$. The label is an in-KB event $e\in\mathcal{E}$ when the mention refers to a KB entry, and is \textit{NIL} when the mention denotes an event instance not covered by $Q$ \cite{liu2026sefel}.

\section{Method}

\begin{figure*}[!t]
    \centering
    \includegraphics[width=1\textwidth]{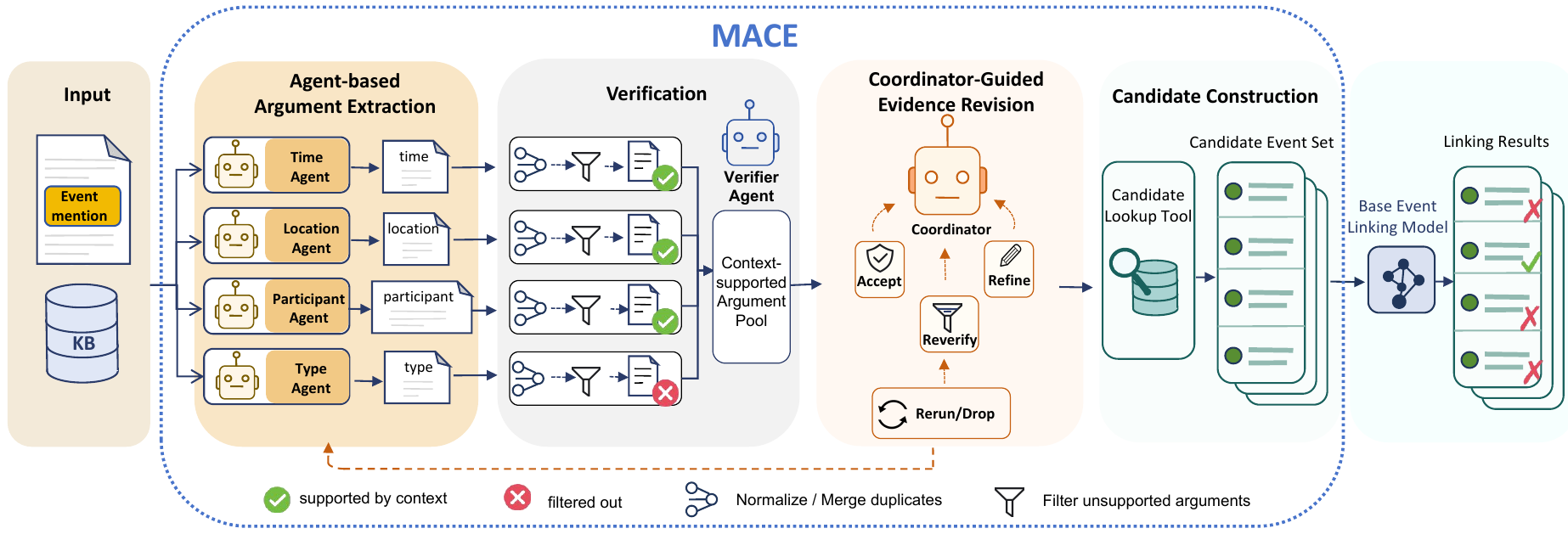}
    \caption{Overview of our framework MACE.}
    \label{framework}
\end{figure*}

The whole process of MACE is shown in Figure~\ref{framework}. 
% MACE is a candidate event acquisition plug-in placed before the event linking model. Next, we will introduce its each part as follows.
% It first uses role-specialized LLM agents to extract and verify time, location, participant, and event-type evidence. A bounded coordinator then sends intermediate queries to a candidate-event lookup tool and uses the returned results to accept, refine, reverify, drop weak evidence, or rerun a role agent. The retained evidence is converted into templated KB lookup inputs and merged into the candidate set $\mathcal{C}_m$, which is passed unchanged to SEFEL, EveLink, or GENRE.

\subsection[Agent-Based Argument Extraction]{Agent-Based Argument Extraction}
Let $\mathcal{P}=\{\text{time},\text{location},\text{participant},\text{type}\}$ denote the evidence-type set. For each evidence type $p \in \mathcal{P}$, an evidence-specialized LLM agent $F_p$ receives $(m,\mathcal{T}_m)$ and returns a raw argument set $\mathcal{A}^{(0)}_{m,p}$. This agent-based design differs from a single generic agent: each agent is instructed to focus on one evidence type required by event linking. Each argument $a \in \mathcal{A}^{(0)}_{m,p}$ is represented as $(p,\operatorname{span}(a),\operatorname{norm}(a),c(a),d(a))$, where $p$ is the evidence type, $\operatorname{span}(a)$ is the extracted text, $\operatorname{norm}(a)$ is a canonicalized form of the span for matching and duplicate removal, $c(a)$ is the agent confidence, and $d(a)$ is the character distance between $\operatorname{span}(a)$ and the event mention. The raw argument pool is
\begin{equation}
\mathcal{A}^{(0)}_m =
\cup_{p\in\mathcal{P}}\mathcal{A}^{(0)}_{m,p}
= \cup_{p\in\mathcal{P}}F_p(m,\mathcal{T}_m).
\end{equation}

\subsection[Verification]{Verification}
Before these arguments are used for candidate event acquisition, MACE applies three ordered operators. $\operatorname{Verify}(\cdot;m,\mathcal{T}_m)$ is a verifier agent that checks whether each argument is supported by the event mention and its context, $\operatorname{Normalize}(\cdot)$ merges duplicate surface forms and normalized values, and $\operatorname{Filter}_{\Delta}(\cdot)$ removes pronominal, contradictory, or distant arguments whose $d(a)$ exceeds the distance threshold $\Delta$. The resulting context-supported argument pool is
\begin{equation}
\begin{aligned}
\mathcal{V}_m &=
\operatorname{Verify}(\mathcal{A}^{(0)}_m;m,\mathcal{T}_m),\\
\mathcal{A}^{(1)}_m &=
\operatorname{Filter}_{\Delta}\big(\operatorname{Normalize}(\mathcal{V}_m)\big).
\end{aligned}
\end{equation}
Here, $\mathcal{V}_m$ denotes the verifier output, and $\mathcal{A}^{(1)}_m$ denotes the context-supported argument pool, namely the verified typed arguments grounded in $\mathcal{T}_m$ after normalization and filtering. This step preserves only typed event evidence that is grounded in the local context and suitable for candidate event acquisition.

\subsection[Coordinator-Guided Evidence Revision]{Coordinator-Guided Evidence Revision}
The context-supported argument pool can still contain true but weakly useful arguments for candidate event acquisition. MACE therefore treats evidence selection as a bounded coordinator loop. At round $t$, the system state $S_t$ stores the verified arguments, selected arguments, templated queries, candidate-event lookup results, and the previous coordinator actions. Given $S_t$, a coordinator agent $G$ chooses an action $u_t$ from five actions: \textsc{accept}, \textsc{refine}, \textsc{reverify}, \textsc{drop}, or \textsc{rerun}. The coordinator uses at most two rounds and its prompt is given in Appendix~\ref{app:prompts}.
\begin{equation}
u_t = G(S_t).
\end{equation}

The coordinator can accept the current evidence set, send the selected arguments to a refinement pass, reapply verification, remove weak spans, or request another extraction pass for a specific evidence type. After the coordinator stops or reaches the maximum number of rounds, MACE applies a salience-aware selector. Let $s(a)$ be the selector score of argument $a$. We use
\begin{equation}
s(a)=w_c c(a)+w_r r(a)+w_n n(a)+w_p p(a).
\end{equation}
Here, $c(a)$ is extractor confidence; $r(a)$ is the role prior; $n(a)$ is the proportion of argument tokens not covered by the mention or previously selected arguments; and $p(a)=1/(1+d(a))$ is proximity. The nonnegative weights $w_c,w_r,w_n,w_p$ sum to one and are specified in the experimental setting. The role-prior values, the total argument budget $L$, the per-type budget $U$, and the distance threshold $\Delta$ are likewise fixed there. Greedy selection uses $\operatorname{Select}_{L,U}$ to rank arguments by $s(a)$, remove duplicate spans or normalized values, enforce the two caps, and return

\begin{equation}
\widehat{\mathcal{A}}_m =
\operatorname{Select}_{L,U}(\mathcal{A}^{(1)}_m;s).
\end{equation}
Here, $\widehat{\mathcal{A}}_m$ denotes the retained argument set used for candidate construction.

\subsection[Candidate Construction]{Candidate Construction}
For candidate event acquisition, MACE organizes $\widehat{\mathcal{A}}_m$ into template inputs. If $\widehat{\mathcal{A}}_m$ is non-empty, let $a^\star$ be the retained argument with the highest selector score $s(a)$, and let $\mathcal{B}_m=\widehat{\mathcal{A}}_m\setminus\{a^\star\}$ be the remaining retained arguments. MACE instantiates inputs with the event mention alone, the event mention with one retained argument, and the event mention with $a^\star$ plus different combinations of arguments from $\mathcal{B}_m$. Let $\mathcal{I}_m$ be the resulting template-input set, and let $D_Q(i)$ be the ranked candidate events returned by the KB dictionary for input $i \in \mathcal{I}_m$. The candidate-event lookup results observed by the coordinator at round $t$ are $\mathcal{F}_t=\{D_Q(i):i\in\mathcal{I}_m\}$, and $\operatorname{Update}(S_t,u_t,\mathcal{F}_t)$ records the returned lookup results together with the chosen action and produces the next state $S_{t+1}$. After the coordinator stops, let $K$ be the maximum number of candidate events retained for each mention. Following the dictionary-based retrieval strategy used in SEFEL~\cite{liu2026sefel}, each input is submitted to the dictionary independently, and the top-ranked results are merged to construct a more comprehensive candidate event set:
\begin{equation}
\mathcal{C}_m =
\operatorname{TopK}_{K}(
\cup_{i\in\mathcal{I}_m}D_Q(i)
)\cup\{\text{NIL}\}.
\end{equation}
Because MACE changes only candidate event acquisition and exposes $\mathcal{C}_m$ to the event linking model, it can be inserted before event linking frameworks such as SEFEL~\cite{liu2026sefel}, EveLink~\cite{yu2023event}, and GENRE~\cite{de2020autoregressive} without modifying their model architectures.

\section{Experiments}
Following prior event linking studies \cite{yu2023event,hsu2024argument,liu2026sefel}, we use accuracy as the main end-to-end metric and evaluate on the Wikipedia and New York Times (NYT) datasets. Wikipedia is an in-KB benchmark built from English Wikipedia hyperlinks, with 66425 training, 16692 validation, and 19267 test mentions. NYT is manually annotated from 2500 lead paragraphs of the New York Times Annotated Corpus, with 769 in-KB and 993 out-of-KB test mentions. MACE uses LLaMA3.1-8B as the LLM backbone. The selector uses $s(a)=w_c c(a)+w_r r(a)+w_n n(a)+w_p p(a)$, with implementation values $(w_c,w_r,w_n,w_p)=(0.30,0.25,0.25,0.20)$; its role priors are 0.85 (participant), 0.65 (time), 0.60 (location), and 0.45 (type). We retain at most $L=5$ arguments in total and $U=2$ arguments per evidence type, use the distance threshold $\Delta=800$ characters and at most $T=2$ coordinator rounds, and keep $K=30$ candidate events.

\subsection{Effectiveness Study} 
We instantiate MACE with three event linking base models: GENRE~\cite{de2020autoregressive}, EveLink~\cite{yu2023event}, and SEFEL~\cite{liu2026sefel}. Table~\ref{tab:main_results} shows that adding the same candidate event acquisition module consistently improves the base models on the event linking benchmarks, indicating that MACE is effective as a plug-in module rather than a component tailored to a single event linking model.

\begin{table}[t]
\centering

\renewcommand{\arraystretch}{1.2}

\setlength{\tabcolsep}{3.5pt}
\resizebox{\columnwidth}{!}{%
\begin{tabular}{lccc|ccc}
\toprule
\multirow{2}{*}{\textbf{\emph{Method}}} & \multicolumn{3}{c|}{\textbf{\emph{Wikipedia Test}}} & \multicolumn{3}{c}{\textbf{\emph{NYT Test}}} \\
\cmidrule(lr){2-4} \cmidrule(lr){5-7}
& \emph{All} & \emph{Verb} & \emph{Noun} & \emph{All} & \emph{Verb} & \emph{Noun} \\
\midrule
GENRE (ICLR 2021) & 26.55 & 13.48 & 39.61 & 23.54 & 22.94 & 24.71 \\
\textbf{GENRE+MACE} & \textbf{43.65} & \textbf{30.09} & \textbf{57.20} & \textbf{42.78} & \textbf{41.76} & \textbf{44.79} \\
EveLink (EACL 2023) & 79.22 & 78.07 & 79.93 & 32.03 & 34.34 & 25.13 \\
\textbf{EveLink+MACE} & \textbf{84.80} & \textbf{86.68} &  \textbf{82.93} & \textbf{67.29} &  \textbf{71.24} & \textbf{55.53} \\

% GPT-4o-mini  & 79.30 & 81.94 & 76.66 & 30.55 & 33.90 & 20.37 \\
SEFEL (AAAI 2026) & 83.64 & 85.99 & 81.29 & 76.90 & 79.61 & 68.85 \\
\textbf{SEFEL+MACE} & \textbf{84.83} & \textbf{87.00} & \textbf{82.67} & \textbf{79.44} & \textbf{81.41} & \textbf{73.59} \\
\bottomrule
\end{tabular}
}

\caption{Performance on the task of event linking. The reported metric is accuracy (ACC, \%). The EveLink Wikipedia value in the K-expansion control is a separately reproduced run; see the accompanying note and Appendix~\ref{app:additional_results}.} 
\label{tab:main_results}
\end{table}

For the SEFEL backbone on Wikipedia, the 1.19 percentage-point gain is statistically supported: SEFEL obtains 83.64\% (standard error 0.27pp; Wilson 95\% CI [83.11, 84.15]) and SEFEL+MACE obtains 84.83\% (standard error 0.26pp; Wilson 95\% CI [84.32, 85.33]). Over 100,000 paired bootstrap samples, the difference has standard deviation 0.17pp and 95\% CI [+0.86,+1.54]; an exact McNemar test gives $p=3.65\times10^{-12}$.

\subsection{Ablation Study}
Table~\ref{tab:mace_ablation} shows that removing any of the three candidate event acquisition components lowers accuracy. Removing agent-based extraction causes the largest drop, especially on NYT, while removing verification or the coordinator loop produces smaller but still substantial declines. The w/o Coordinator variant retains the final salience selector, so its result isolates the contribution of coordinator-guided revision. Overall, the results suggest that evidence-specialized argument acquisition is the main source of improvement, with verification and coordinator-guided revision providing complementary benefits.

The role-extension comparison is provided in Appendix~\ref{app:additional_results}; they are auxiliary analyses and do not change the default four-role configuration used in Table~\ref{tab:mace_ablation}.

\begin{table}[!htbp]
\centering
\small
\setlength{\tabcolsep}{4pt}
\renewcommand{\arraystretch}{1.16}
\resizebox{0.8\columnwidth}{!}{%
\begin{tabular}{lcc}
\toprule
\multirow{2}{*}{\textbf{\emph{Ablation}}} & \multicolumn{1}{c|}{\textbf{\emph{Wikipedia Test}}} & \multicolumn{1}{c}{\textbf{\emph{NYT Test}}} \\
\cmidrule(lr){2-2} \cmidrule(lr){3-3}
& \emph{All} & \emph{All} \\
\midrule
SEFEL + MACE & 84.83 & 79.44 \\
w/o Agent Extraction & 82.22 & 56.53 \\
w/o Verification & 84.11 & 76.92 \\
w/o Coordinator & 83.76 & 75.97 \\

\bottomrule
\end{tabular}
}
\caption{Ablation study results. The w/o Agent Extraction, w/o Verification, and w/o Coordinator rows remove the corresponding extraction, verification, and coordinator-loop components while keeping the downstream SEFEL linker fixed.
% of the MACE candidate event acquisition module with the fixed SEFEL event linking model. All variants are evaluated on the same processed Wikipedia and NYT test sets, so only candidate event acquisition changes across rows.
}
\label{tab:mace_ablation}
\end{table}

To separate MACE's evidence refinement from a generic increase in candidate-set size, we evaluate naive candidate expansion on Wikipedia. The control results and the interpretation are given in Appendix~\ref{app:additional_results}; increasing EveLink from $K=30$ to $K=100$ improves accuracy by 3.65pp, but remains 1.93pp below EveLink+MACE at $K=30$. For SEFEL, the same expansion changes accuracy by only 0.06pp, while MACE yields a 1.19pp gain at $K=30$. These comparisons indicate that MACE contributes informative evidence beyond candidate-set size alone.

\subsection{Efficiency Study}
Table~\ref{tab:efficiency} provides the efficiency comparison for the three transfer base models. The table contrasts each original event linking models with the same model augmented by MACE candidate event acquisition under the same evaluation setting. Under this setting, adding MACE reduces the measured runtime for GENRE and EveLink and leaves SEFEL unchanged, indicating that the plug-in improves or preserves efficiency across the evaluated event linking model.

\begin{table}[!htbp]
\centering
\small
\setlength{\tabcolsep}{3.5pt}
\renewcommand{\arraystretch}{1.16}
\resizebox{0.8\columnwidth}{!}{%
\begin{tabular}{lcccc}
\toprule
\multirow{2}{*}{\textbf{\emph{Base Model}}} & \multicolumn{2}{c}{\textbf{\emph{Wikipedia Test}}} & \multicolumn{2}{c}{\textbf{\emph{NYT Test}}} \\
\cmidrule(lr){2-3} \cmidrule(lr){4-5}
& \emph{Original} & \emph{+MACE} & \emph{Original} & \emph{+MACE} \\
\midrule
GENRE & 161325 & \textbf{107019} & 6927 & \textbf{4595} \\
EveLink & 24614 & \textbf{17501} & 2352 & \textbf{1731} \\
SEFEL & \textbf{446} & \textbf{446} & \textbf{59} & \textbf{59} \\
\bottomrule
\end{tabular}
}
\caption{Efficiency study results. Runtime is measured in seconds. The reported values are downstream linker runtime with pre-cached MACE outputs; online MACE latency is analyzed in the text and Appendix~\ref{app:additional_results}.
% between the original event linkers and MACE-enhanced linkers. Runtime is reported as downstream linker time after cached candidate acquisition under the same hardware configuration.
}
\label{tab:efficiency}
\end{table}

The cached setting in Table~\ref{tab:efficiency} excludes the LLM calls used to create MACE outputs. In an online deployment, a reconstruction on 200 mentions with LLaMA3.1-8B and the two-round coordinator limit estimates an additional 12.23 seconds per Wikipedia mention and 9.62 seconds per NYT mention. This cost can be amortized in batch preprocessing, whereas latency-sensitive applications may need caching, a faster backend, or selective MACE invocation for ambiguous mentions. The stage-level estimates are listed in Appendix~\ref{app:additional_results}.

\subsection{Qualitative Analysis}
We illustrate the evidence-state transitions for the Wikipedia mention ``coming war,'' whose gold KB event is World War II. The example shows how verification removes unsupported evidence and how coordinator feedback leads to a more retrieval-useful argument set.

\begin{table}[t]
\centering
\scriptsize
\setlength{\tabcolsep}{4pt}
{
\begin{tabular}{p{0.24\columnwidth}p{0.70\columnwidth}}
\toprule
\textbf{Stage} & \textbf{Evidence state/action} \\
\midrule
Raw role outputs & Time: ``1935,'' ``7 April 1939''; participant: ``Hughes,'' ``Billy Hughes''; additional cues: ``UAP,'' ``Australia,'' and ``Australia and the War Today.'' \\
Verified/filtered evidence & Removes the unrelated date ``7 April 1939'' and normalizes the overlapping ``Hughes'' mentions. \\
Coordinator action & \textsc{REFINE}, because the provisional candidate feedback remains broad. \\
Final retained evidence & ``1935,'' ``Billy Hughes,'' and ``Australia and the War Today.'' \\
Candidate change & The original list ranks Battle for Australia first and omits World War II from its 30 slots; MACE ranks World War II first. \\
\bottomrule
\end{tabular}}
\caption{Case of evidence refinement for the ``coming war'' mention.}
\label{tab:case_study}
\end{table}

This case is illustrative: the observed gain comes from removing a misleading date, consolidating duplicate participants, and retaining evidence that produces a more useful candidate query.

\section{Related Work}

% Early work on event linking grounded event references in news archives \cite{nothman2012event}. 
More recent work formally defines event linking as aligning textual event mentions with corresponding event nodes in a structured KB such as Wikipedia or Wikidata \cite{yu2023event,hsu2024argument,liu2026sefel,liu2023low,shen2018predicting}. These systems mostly adopt a two-stage retrieve-and-rank framework: a bi-encoder retrieves candidate events, and a computationally intensive cross-encoder performs fine-grained ranking over the retrieved candidate events. Their main differences often lie in how they perform candidate event retrieval: named-entity augmentation in EveLink~\cite{yu2023event}, or argument-aware retrieval in ArgEvelink~\cite{hsu2024argument}. SEFEL~\cite{liu2026sefel} reduces the reranking cost by replacing the cross-encoder stage with a fast argument-aware bi-encoder. MACE is orthogonal to these event linking frameworks because it improves candidate event acquisition before the event linking model receives its candidate event set.

Event argument extraction (EAE) identifies the arguments associated with an event mention. More recent work has moved toward document-level memory and retrieval \cite{liu2025compressive,lin-etal-2025-generation}, hybrid selection-generation decoding \cite{ding2025fusion}, and LLM-supported document-level augmentation \cite{gatto-etal-2025-document}. LLM-based prompting also remains a practical route when annotated supervision is limited \cite{chen2024large}. Our setting is narrower than general EAE: we do not need exhaustive event-argument annotation, but a small set of arguments that helps event disambiguation during candidate event acquisition. MACE therefore emphasizes evidence-specialized extraction, explicit verification, and salience ranking before these arguments are used for candidate event acquisition.

Recent surveys describe a broader shift from single-agent prompting to coordinated LLM systems that split work by evidence type, route, or consensus \cite{xi2025rise}. This trend has already reached structured information extraction. MMD-ERE~\cite{guan-etal-2025-mmd-ere} applies agent debate to event relation extraction. These studies show that specialized agents can improve extraction quality, but they do not address candidate event acquisition for event linking. MACE follows an agent-specialized pipeline principle: acquiring typed event arguments, verifying them before use, and keeping only the arguments most useful for candidate event acquisition and linking.

\section{Conclusion}
In this paper, we present MACE, a candidate event acquisition method that refines event structure before linking. Extensive experiments show that MACE can improve many event linking models on both Wikipedia and NYT. 
% These results confirm that event-structure refinement can strengthen event linking through reusable candidate event acquisition.

\section{Acknowledgments}
The work was partially supported by the National Key Research and Development Program of China (No. 2024YFF0617702); the National Natural Science Foundation of China (Nos. U22A2025, 62402097, 62232007, and U23A20309); the Joint Funds of the Natural Science Foundation of Liaoning Province (No. 2023-BSBA-132); the 111 Project (No. B16009); and the Fundamental Research Funds for the Central Universities (No. N2417007). 

\section{Limitations}
MACE introduces an LLM-based candidate event acquisition stage before the event linking model. In our experiments, this stage is executed as cached preprocessing, while the efficiency comparison reports the downstream event linking model runtime under fixed candidate sets; fully online use would introduce additional API latency and cost. In addition, our evaluation studies transfer mainly with SEFEL, EveLink, and GENRE on Wikipedia and NYT. Broader validation across more domains, knowledge bases, and event linking architectures remains necessary. Finally, MACE focuses on time, location, participant, and event-type evidence; domains whose disambiguation depends on finer-grained event semantics may require extending the evidence-specialized agents.

\bibliography{cas-refs}

\clearpage
\appendix

\section{Additional Results and Analyses}
\label{app:additional_results}

\subsection{Candidate-Set Expansion Control}
\begin{table}[h]
\centering
\scriptsize

{
\begin{tabular}{lcc}
\toprule
Acquisition setting & $K$ & Wikipedia ACC (\%) \\
\midrule
EveLink & 30 & 79.22 \\
EveLink & 50 & 80.27 \\
EveLink & 100 & 82.87 \\
EveLink + MACE & 30 & 84.80 \\
SEFEL & 30 & 83.64 \\
SEFEL & 50 & 83.68 \\
SEFEL & 100 & 83.70 \\
SEFEL + MACE & 30 & 84.83 \\
\bottomrule
\end{tabular}}
\caption{Naive candidate-set expansion control.}
\label{tab:k_control}
\end{table}

Increasing EveLink from $K=30$ to $K=100$ gives a 3.65pp gain, but the $K=100$ result remains 1.93pp below EveLink+MACE at $K=30$. For SEFEL, increasing $K$ changes accuracy by only 0.06pp, whereas MACE gives a 1.19pp gain at $K=30$. Thus, the MACE improvement is not explained by candidate-set size alone.

\subsection{Latency Breakdown}
\begin{table}[h]
\centering
\scriptsize
\setlength{\tabcolsep}{2.6pt}
{
\begin{tabular}{lcc}
\toprule
MACE stage & Wikipedia (s) & NYT (s) \\
\midrule
Time/location/participant/type extraction & 2.67 & 2.46 \\
Contextual verification & 3.16 & 2.91 \\
Deterministic filtering and lookup & 0.12 & 0.10 \\
First coordinator decision & 1.75 & 1.61 \\
Conditional refinement & 1.24 & 0.29 \\
Second coordinator decision & 1.16 & 0.31 \\
Salience selection & 2.12 & 1.95 \\
Expected MACE front-end total & 12.23 & 9.62 \\
\bottomrule
\end{tabular}
}
\caption{Engineering estimate of online MACE latency per mention, reconstructed from 200 samples per dataset with LLaMA3.1-8B and a two-round coordinator limit.}
\label{tab:latency_breakdown}
\end{table}

The estimates include the LLM calls and deterministic lookup stages that are excluded from the cached downstream runtime in Table~\ref{tab:efficiency}. They quantify latency rather than a monetary API price; the latter depends on the deployment backend and request batching.

\subsection{Role Extensibility}
\begin{table}[h]
\centering
\scriptsize
\setlength{\tabcolsep}{4pt}
{
\begin{tabular}{lcc}
\toprule
Role configuration & ACC (\%) & Recall@5 (\%) \\
\midrule
Four roles (default) & 86.67 & 93.33 \\
Four roles + Instrument + Purpose + Cause & 86.67 & 93.33 \\
\bottomrule
\end{tabular}}
\caption{Role extensibility on a randomly sampled 5\% of the Wikipedia test set using the same SEFEL checkpoint.}
\label{tab:role_extension}
\end{table}

The additional roles can be inserted without changing the pipeline interface, but they do not improve this sampled subset. Clause-form Cause and Purpose evidence is often mapped to NIL when treated as standalone entity spans, motivating the four broadly applicable default roles.

\subsection{Extraction-Front-End Comparison}
The following comparison addresses the distinction between event-extraction outputs and linking-oriented evidence acquisition. All settings use the same downstream SEFEL linker; they differ only in the front end that supplies evidence or candidates.

\begin{table}[h]
\centering
\scriptsize
\setlength{\tabcolsep}{2.2pt}
{
\resizebox{\columnwidth}{!}{%
\begin{tabular}{lcccccc}
\toprule
\multirow{2}{*}{Method} & \multicolumn{3}{c|}{Wikipedia} & \multicolumn{3}{c}{NYT} \\
\cmidrule(lr){2-4} \cmidrule(lr){5-7}
 & All & Verb & Noun & All & Verb & Noun \\
\midrule
SEFEL\_Uni+Tag & 83.28 & 85.27 & 80.85 & 62.43 & 66.72 & 49.66 \\
SEFEL & 83.64 & 85.99 & 81.29 & 76.90 & 79.61 & 68.85 \\
SEFEL+MACE & 84.83 & 87.00 & 82.67 & 79.44 & 81.41 & 73.59 \\
\bottomrule
\end{tabular}}
}
\caption{Indirect comparison of traditional extraction, the original SEFEL front end, and MACE under the same downstream linker.}
\label{tab:front_end_comparison}
\end{table}

For the NYT mention ``proceedings'' in ``investor concern about impeachment proceedings against President Clinton hurt the dollar,'' the gold event is Impeachment of Bill Clinton. Table~\ref{tab:front_end_case} details how the evidence and induced candidates differ between the traditional and LLM-based front ends.

\begin{table*}[t]
\centering
\scriptsize
\setlength{\tabcolsep}{4pt}
{
\begin{tabular}{p{0.22\textwidth}p{0.35\textwidth}p{0.35\textwidth}}
\toprule
Stage & Traditional UniST + TagPrime & LLM-based SEFEL front end \\
\midrule
Extracted evidence & Long span ``Weakness in the stock market ... President Clinton''; ``to 116 42 yen'' (duplicated); ``the dollar'' & Participant: Clinton; type: ``impeachment proceedings''\\
Candidates induced by the extracted evidence & NIL; Conference proceeding; Dollar; Shadrake v Attorney-General; Proceedings (magazine) & Impeachment of Bill Clinton; Shadrake v Attorney-General; Impeachment inquiry against Donald Trump \\
Gold event present in this evidence-derived candidate list & No & Yes; it is the highest-scored entry in the stored list \\
\bottomrule
\end{tabular}}
\caption{Case study of evidence quality for the NYT mention ``proceedings.''}
\label{tab:front_end_case}
\end{table*}

The traditional extractor's spans may be locally reasonable under its extraction formulation, but they are less discriminative for event linking. The LLM-based front end recovers participant, event-type evidence that makes the gold event retrievable.

\section{Prompt and Implementation Details}

\label{app:prompts}
For reproducibility, we include the full prompt templates and decoding constraints used by MACE. The code repository is available at \href{https://github.com/Princess-nil/MACE}{the code repository}.

\subsection{Prompt Inventory}
The reported implementation uses the following prompt family: \texttt{TimeAgent}, \texttt{LocationAgent}, \texttt{ParticipantAgent}, \texttt{EventTypeAgent}, \texttt{VerifyAgent}, \texttt{RefineAgent}, and \texttt{CoordinatorAgent}. The extraction-style agents share a JSON contract, while the coordinator uses a separate action schema.

\begin{promptbox}{Shared Extraction Contract}
Return ONLY valid JSON (no markdown fences, no commentary). Schema:
{
    "arguments": [
        {
            "type": "{expected_type}",
            "span_text": string,
            "start": integer,
            "ln": integer,
            "confidence": number
        }
    ]
}
Rules:
- span_text MUST be an EXACT substring of the Text (case-sensitive, character-perfect).
- start is the 0-based character offset of span_text in Text.
- ln is the character length of span_text (must equal len(span_text)).
- confidence is a float in [0, 1].
- Extract ONLY arguments relevant to the target event mention.
- If none found, return {"arguments": []}.
\end{promptbox}

\begin{promptbox}{TimeAgent}
You are TimeAgent. Your sole task is to extract TIME expressions that are directly relevant to the target event mention in the given text. You are precise and conservative -- only extract temporal expressions that describe WHEN the event happened, its duration, or a closely related time frame. Do NOT extract times for unrelated events mentioned in the same passage.

Text:
{text}

Event mention: "{mention}" (start={mention_start}, end={mention_end})

Task: extract TIME arguments for this event.

Return ONLY valid JSON (no markdown fences, no commentary). Schema:
{
    "arguments": [
        {
            "type": "{expected_type}",
            "span_text": string,
            "start": integer,
            "ln": integer,
            "confidence": number
        }
    ]
}
Rules:
- span_text MUST be an EXACT substring of the Text (case-sensitive, character-perfect).
- start is the 0-based character offset of span_text in Text.
- ln is the character length of span_text (must equal len(span_text)).
- confidence is a float in [0, 1].
- Extract ONLY arguments relevant to the target event mention.
- Prefer short, precise spans over long phrases.
- If none found, return {"arguments": []}.
\end{promptbox}

\begin{promptbox}{LocationAgent}
You are LocationAgent. Your sole task is to extract LOCATION/PLACE spans that are directly relevant to the target event mention. Extract specific place names (cities, countries, venues, regions) tied to where the event occurred or is situated. Avoid overly broad locations unless the text explicitly ties them to the event.

Text:
{text}

Event mention: "{mention}" (start={mention_start}, end={mention_end})

Task: extract LOCATION arguments for this event.

Return ONLY valid JSON (no markdown fences, no commentary). Schema:
{
    "arguments": [
        {
            "type": "{expected_type}",
            "span_text": string,
            "start": integer,
            "ln": integer,
            "confidence": number
        }
    ]
}
Rules:
- span_text MUST be an EXACT substring of the Text (case-sensitive, character-perfect).
- start is the 0-based character offset of span_text in Text.
- ln is the character length of span_text (must equal len(span_text)).
- confidence is a float in [0, 1].
- Extract ONLY arguments relevant to the target event mention.
- Prefer short, precise spans over long phrases.
- If none found, return {"arguments": []}.
\end{promptbox}

\begin{promptbox}{ParticipantAgent}
You are ParticipantAgent. Your sole task is to extract PARTICIPANT spans (people, organizations, groups, key objects) that are directly involved in the target event. Prefer named entities explicitly mentioned in the same or adjacent sentence as the mention. Do NOT extract pronouns or generic references.

Text:
{text}

Event mention: "{mention}" (start={mention_start}, end={mention_end})

Task: extract PARTICIPANT arguments for this event.

Return ONLY valid JSON (no markdown fences, no commentary). Schema:
{
    "arguments": [
        {
            "type": "{expected_type}",
            "span_text": string,
            "start": integer,
            "ln": integer,
            "confidence": number
        }
    ]
}
Rules:
- span_text MUST be an EXACT substring of the Text (case-sensitive, character-perfect).
- start is the 0-based character offset of span_text in Text.
- ln is the character length of span_text (must equal len(span_text)).
- confidence is a float in [0, 1].
- Extract ONLY arguments relevant to the target event mention.
- Prefer short, precise spans over long phrases.
- If none found, return {"arguments": []}.
\end{promptbox}

\begin{promptbox}{EventTypeAgent}
You are EventTypeAgent. Your sole task is to extract short descriptors of the EVENT CATEGORY or nature for the target event (e.g., 'war', 'election', 'assassination', 'merger', 'tournament'). Only extract spans that appear verbatim in the text. If no supported event-type descriptor is present, return an empty list.

Text:
{text}

Event mention: "{mention}" (start={mention_start}, end={mention_end})

Task: extract EVENT-TYPE arguments for this event.

Return ONLY valid JSON (no markdown fences):
{
    "arguments": [
        {
            "type": "type",
            "span_text": string,
            "start": integer,
            "ln": integer,
            "confidence": number
        }
    ]
}
Rules:
- span_text MUST be an EXACT substring of Text.
- start is 0-based char offset, ln = len(span_text).
- Use "type" for event-category descriptors.
- If none found, return {"arguments": []}.
\end{promptbox}

\begin{promptbox}{VerifyAgent}
You are VerifyAgent. You receive candidate event arguments extracted by other agents. Your job is to:
1) Remove arguments NOT supported by the text or irrelevant to the mention.
2) Remove duplicates (same or overlapping spans with same semantics).
3) Resolve conflicts (e.g., contradictory times, redundant locations).
4) Keep only arguments from sentences containing or adjacent to the mention.
Be conservative: when in doubt, keep the argument.

Text:
{text}

Event mention: "{mention}" (start={mention_start}, end={mention_end})

Candidate arguments:
{candidates_json}

Task: output the filtered, deduplicated arguments.

Return ONLY valid JSON:
{
    "arguments": [
        {
            "type": string,
            "span_text": string,
            "start": integer,
            "ln": integer,
            "confidence": number
        }
    ]
}
Preserve the original type values (time/location/participant/type).
If all candidates are valid, return them unchanged.
\end{promptbox}

\begin{promptbox}{RefineAgent}
You are RefineAgent. You perform a second-pass refinement for event linking. You receive the text, the target event mention, a set of verified arguments, and provisional retrieval queries built from those arguments. Your job is to keep the arguments that best help distinguish this event from other plausible candidates. Remove generic, weak, or redundant arguments. Prefer arguments that would sharpen retrieval, such as event-type descriptors, key participants, distinctive locations, or precise times.

Text:
{text}

Event mention: "{mention}" (start={mention_start}, end={mention_end})

Verified arguments:
{arguments_json}

Provisional queries:
{queries_json}

Task: return the refined arguments that should be kept after the second pass.

Return ONLY valid JSON:
{
    "arguments": [
        {
            "type": string,
            "span_text": string,
            "start": integer,
            "ln": integer,
            "confidence": number
        }
    ]
}
Preserve the original type values when an argument is kept. If no change is needed, return the most useful subset unchanged.
\end{promptbox}

\begin{promptbox}{CoordinatorAgent}
You are CoordinatorAgent for event linking candidate acquisition. You observe extracted arguments, candidate-event lookup feedback, and the action history. Choose one next action. Available actions are: accept, refine, reverify, drop, rerun. Use refine when candidate lookup is too broad or noisy; use reverify when arguments conflict or look unsupported; use drop when weak arguments make queries too broad; use rerun only when one evidence type is missing and likely needed. Return only JSON.

Observation:
{
  "round": round_id,
  "verified_arguments": [...],
  "selected_arguments": [...],
  "queries": [...],
  "candidate_feedback": [...],
  "history": [...]
}

Return JSON with schema:
{
  "action": "accept|refine|reverify|drop|rerun",
  "reason": string,
  "target_type": "time|location|participant|type|null",
  "drop_spans": [string]
}
\end{promptbox}

\subsection{Output Format and Parameters}
 \begin{table}
\centering
\scriptsize
\setlength{\tabcolsep}{2.1pt}
\renewcommand{\arraystretch}{1.12}
\begin{tabular}{p{0.30\columnwidth}p{0.61\columnwidth}}
\toprule
\textbf{Item} & \textbf{Value} \\
\midrule
\texttt{arguments} & Extraction, verification, and refinement modules return a JSON list of spans with \texttt{type}, \texttt{span\_text}, \texttt{start}, \texttt{ln}, and \texttt{confidence}. \\
\texttt{action} & The coordinator returns one of \texttt{accept}, \texttt{refine}, \texttt{reverify}, \texttt{drop}, or \texttt{rerun}. \\
\texttt{temperature} & All LLM modules use \texttt{0.0}. \\
\texttt{max\_rounds} & The coordinator loop uses at most \texttt{2} rounds. \\
\texttt{lookup\_topk} & Candidate lookup keeps the top \texttt{10} titles for each provisional query. \\
\texttt{L, U} & Argument selection keeps at most \texttt{L=5} arguments total and \texttt{U=2} arguments per evidence type. \\
\texttt{K} & Final candidate construction keeps \texttt{K=30} candidate events before adding \textit{NIL}. \\
\texttt{backbone} & The reported implementation uses \texttt{LLaMA3.1-8B} as the LLM backbone. \\
\bottomrule
\end{tabular}
\caption{Output format and reproducibility parameters used by MACE.}
\label{tab:appendix_formats}
\end{table}

\begin{promptbox}{Candidate Lookup Templates}
For each mention, MACE forms three query families: the mention alone, the mention paired with each retained argument, and the mention paired with the highest-scoring retained argument plus additional retained arguments. Each query is submitted independently to the KB dictionary, and duplicate returned events are merged before top-$K$ truncation. The final candidate set is augmented with NIL and passed unchanged to the downstream event linker.
\end{promptbox}

\end{document}